\documentclass[runningheads]{llncs}
\usepackage{graphicx}
\usepackage{multirow}
\usepackage{eqnarray}
\usepackage{indentfirst}
\usepackage{color}
\usepackage{float}
\usepackage{subfigure}
\usepackage{diagbox}
\usepackage{amsmath}

\begin{document}
\title{Hierarchical multi-class segmentation of glioma images using networks with multi-level activation function}

\author{Xiaobin Hu\inst{1,2}\and
Hongwei Li\inst{1}\and Yu Zhao\inst{1}\and Chao Dong\inst{1}\and  Bjoern H. Menze\inst{1}\and Marie Piraud\inst{1}}
\authorrunning{X.B.Hu et al.}
\institute{Department of computer science, Technische Universit\"{a}t M\"{u}nchen,\\Munich, Germany \\
\and \email{xiaobin.hu@tum.de}\\
}


%
\maketitle             
\begin{abstract}
 For many segmentation tasks, especially for the biomedical image, the topological prior is vital information which is useful to exploit. The containment/nesting is a typical inter-class geometric relationship. In the MICCAI Brain tumor segmentation challenge, with its three hierarchically nested classes 'whole tumor', 'tumor core', 'active tumor', the nested classes relationship is introduced into the 3D-residual-Unet architecture. The network comprises a context aggregation pathway and a localization pathway, which encodes increasingly abstract representation of the input as going deeper into the network, and then recombines these representations with shallower features to precisely localize the interest domain via a localization path. The nested-class-prior is combined by proposing the multi-class activation function and its corresponding loss function. The model is trained on the training dataset of Brats2018, and 20\% of the dataset is regarded as the validation dataset to determine parameters. When the parameters are fixed, we retrain the model on the whole training dataset. The performance achieved on the validation leaderboard is 86\%, 77\% and 72\% Dice scores for the whole tumor, enhancing tumor and tumor core classes without relying on ensembles or complicated post-processing steps. Based on the same start-of-the-art network architecture, the accuracy of nested-class (enhancing tumor) is reasonably improved from 69\% to 72\% compared with the traditional Softmax-based method which blind to topological prior.
\end{abstract}

\section{Introduction}
Glioma are the most common family of brain tumors, and forms some of highest-mortality and economically costly diseases of brain cancer \cite{Davis,Hanif,Birbrair}. The diagnosed method is highly relayed on manual segmentation and analysis of multi-modal MRI scans by bio-medical experts. Nevertheless, this diagnosed way is severely limited by the labor-intensive character of the manual segmentation process and disagreement or mistakes between manual segmentation. Consequently, there exists a great need for a fast and robust automated segmentation algorithm. Convolutional neural networks (CNNs) have been verified to be extremely effective for a variety of semantic segmentation tasks \cite{Gu}.

While CNN segmentation algorithms are abundant in biomedical imaging, only very few make use of nested-topological prior information. Among the few that do \cite{Nosrati,BenTaieb,Christ,Fidon,Bauer,Alberts,Liu}, we find three different approaches. First, the use of cascaded algorithms where the network consists of successive segmentation networks. Second, the information on the nested-classes is incorporated into the loss function, imposing penalties on solutions that do not respect the nested geometry relations. Third, Markov random fields are used to formalizing class relationship in the post-processing of the network output. Here, we make use of a new activation function \cite{Marie} that is directly implementing class hierarchy in the network training and generalize it to 3 nested classes. For the glioma labels we assume that active tumor regions are always contained in the tumor core which is surrounded by the tumor edema, resulting in a hierarchical three-class model. In sharp contrast with nested-class method, the softmax-based method of multi-class ignores the geometric prior between different classes, and assumes the classes are mutually-exclusive, meaning one pixel cannot belong to different classes at the same time, which absolutely discards the topological information and sometimes leads the unreasonable segmentation results. The comparison of Dice score criteria between two different methods is implemented and it obviously indicates the nested-class method achieves higher accuracy than the softmax-based method, especially for the internal-classes.

In the following, we introduce a brief overview of start-of-the-art 3D-residual U-net architecture and multi-class-nested activation and loss function. We then propose and evaluate our model architectures for Brats tumor segmentation. Finally, we implement the comparison between two main avenues and illustrate the multi-level activation performs better especially in the inter-class. 

\section{Methodology}
\subsection{Network Architecture}
\begin{figure}[!htb]
\centering
\includegraphics[width=12cm]{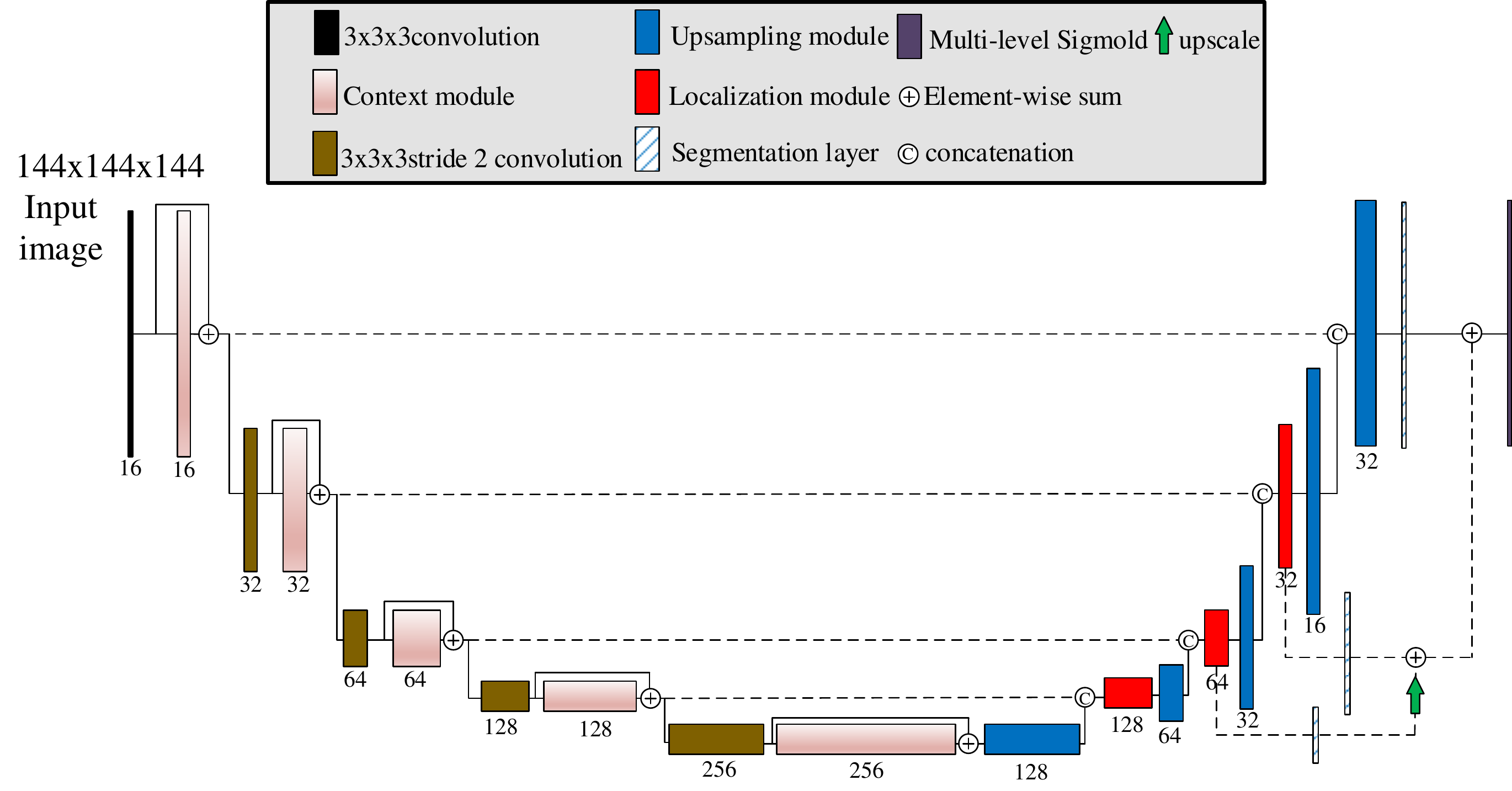}
\caption{Network architecture from \cite{Fabian}: Context pathway (left) aggregates high level information;Localization pathway (right) localizes precisely }\label{fig:fig2}
\end{figure}

The nested-classes relationship between different labels are shown in Fig.2. 
The general network structure shown in Fig.1 is stemming from the previously used glioma segmentation network by Isensee \cite{Fabian} to process large 3D input blocks of 144x144x144 voxels. The original network is inspired by the U-net \cite{Olaf} which allows the network to intrinsically recombine different scales throughout the entire network. This vertical depth is set as 5, which balances between the spatial resolution and feature representations. The context module is a pre-activation residual block, and is connected by 3x3x3 convolutions with input stride 2. The purpose of the localization pathway is to extract features from the lower levels of the network and transform them to a high spatial resolution by means of a simple upscale technology. The upsampled features and its corresponding level of the context aggregation feature are recombined via concatenation. Furthermore, the localization module, consisting of a 3x3x3 convolution followed by a 1x1x1 convolution, is designed to gather these features.
\begin{figure}[!htb]
\centering
\includegraphics[width=6cm]{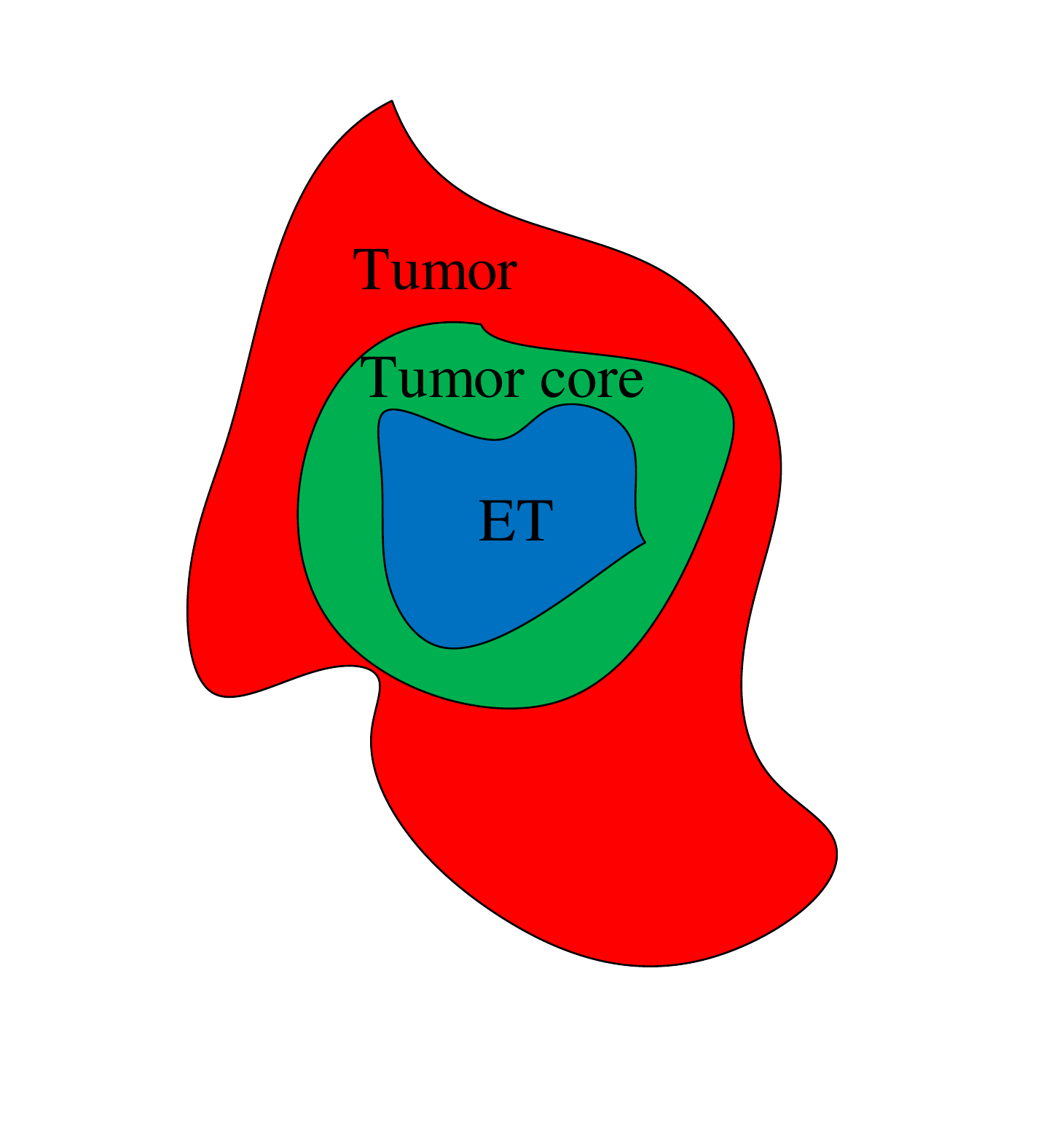}
\caption{Schematic description of the nesting of classes in the BRATS challenge, which respects the following hierarchy: Enhancing Tumor (ET) $\in$Tumor core $\in$ Tumor }\label{fig:fig1}
\end{figure}

The deep supervision is introduced in the localization pathway by integrating segmentation layers at different levels of the network and combining them via elementwise summation to form the final network output. The output activation layer is multi-level Sigmoid layer instead of softmax layer in the Isensee's network which converting the multi-class problem to binary ones. Intrinsically, the multi-level activation is the assemble of multi-sigmoid function and then straightforwardly maps to multi-class segmentation incorporating the topological prior. Consequently, it overcomes the softmax-based method's shortcoming which is blind to the geometric prior.

\subsection{Crop preprocessing}
For 3D network architecture, the larger patch size of training dataset contains more continuous context knowledge and localization information which are beneficial to improve the segmentation accuracy. In order to acquire to the larger cube size patch of 3D image, the valuable knowledge in the MRI is extracted as much as possible while the meaningless information is cropped. Then the crop processing is implemented, and the maximum size of cube patch is selected as [144,144,144].  

The crop preprocessing equation is defined as:

\begin{equation} \label{eq2}
\begin{split}
&array=[a_{min}-(b_{size}-a)/2:a_{min}+(b_{size}+a)/2]\\
&a=a_{max}-a_{min}
\end{split}
\end{equation}
where $a_{min}$ and $a_{max}$ are the min and max non-zero information index of MRI image, and $a$ represents the length of non-zero information.$b_{size}$ is the cube patch size and selected as 144.

The index is recorded and used in the image post-processing stage to recovery back to the original shape [155,240,240].
However, a little of meaningful information which exceeds the cube patch size 144 is unavoidably ignored and have little effect on the segmentation result. In order to equally compare the softmax-based with the multi-level method, no data augmentation operation is used in the stage of image pre-procssing.

\subsection{Multi-level method}

\begin{figure}[!htb]
\centering
\includegraphics[width=8.5cm]{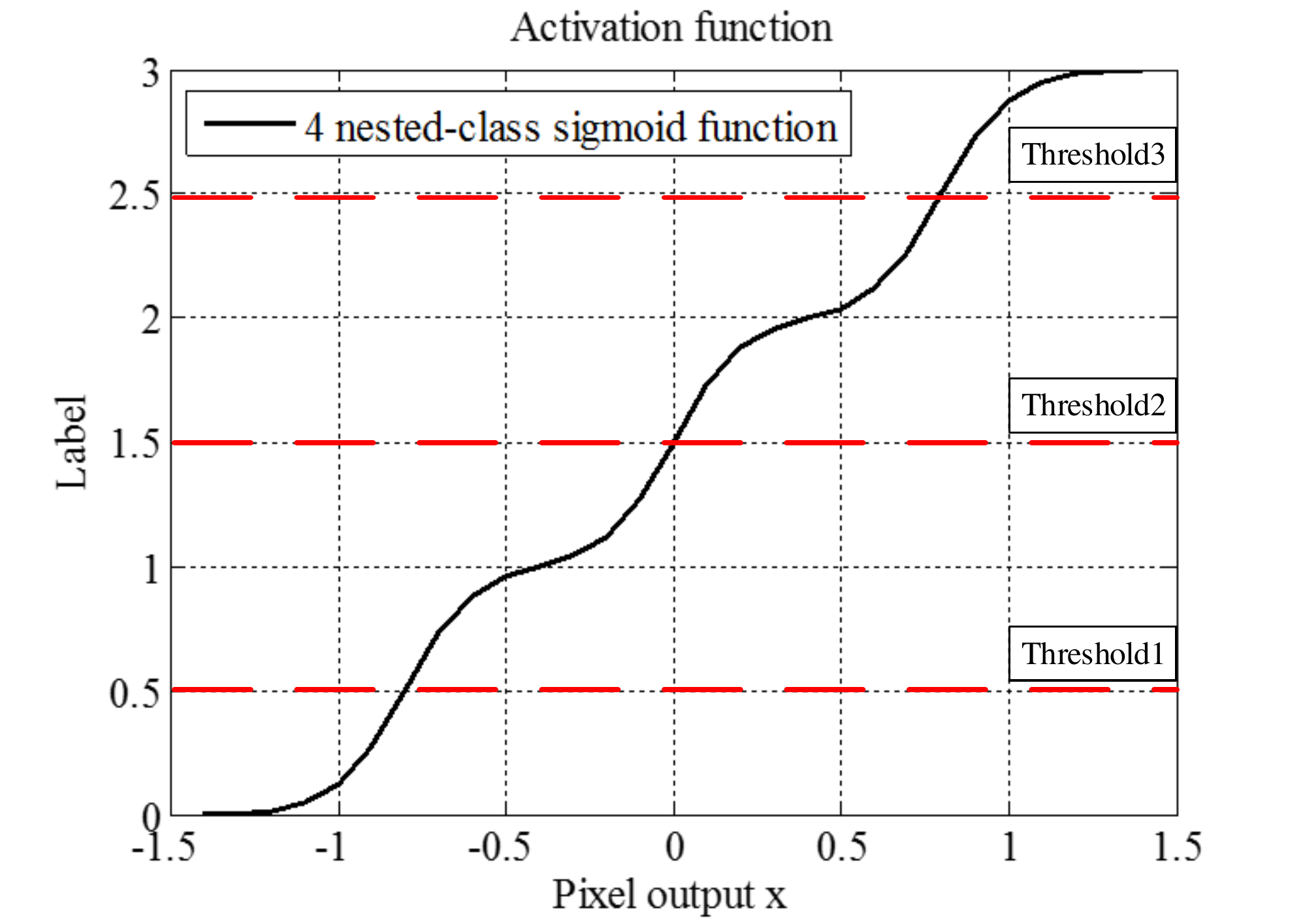}
\caption{Multi-class activation function, Eq.(1) with m+1=4, h=0.8 and k=10 }\label{fig:fig3}
\end{figure}

Here, we use one output channel and a multi-class-nested activation function, as first proposed in \cite{Marie}.The multi-level method is inspired by continuous regression, and thereby generalizing logistic regression to hierarchically-nested classes. It is shown in Fig.3 and defined as 
\begin{equation} \label{eq1}
\begin{split}
a(x)=\sum^m_{n=1}\sigma(k[x+h(n-\frac{m+1}{2})])
\end{split}
\end{equation}

Where $\sigma$ is the sigmoid function, k is the steepness and h is the spacing between consecutive Sigmoids. For Brain tumor segmentation challenge 4-classes nested label case, we have m+1=4, and we take h=0.5 and steepness=10.
The corresponding loss function, called Modified Cross-Entropy (MCE) in \cite{Marie}, is defined as 
\begin{equation} \label{eq2}
\begin{split}
L_{MCE} = - \frac{1}{N_{tot}}\sum_{pixel\,i} \sum_{classes\,c}{y^c_i w^c}log(P^c [a(x_i)])
\end{split}
\end{equation}
where $w^c$ is the weight of corresponding label,which we take as$w^{c\alpha}$($w^{c\alpha}={(\frac{N_{tot}}{N_c}})^{\alpha}$), where $N_{tot}$is the sum number of pixels,$N_{c}$ the number of pixels in each class, and where $y^c=1$ for the ground-truth label c of pixel i and $y^c=0$ otherwise. Furthermore, the mapping function $P^c$ is defined as

\begin{equation} \label{eq3}
\begin{split}
& P^{c=0}(a)=1-a/3 \\
& P^{c=1}(a)=a\Theta(1-a)+(3-a)/2\Theta(a-1) \\
& P^{c=2}(a)=a/2\Theta(2-a)+(3-a)\Theta(a-2) \\ 
& P^{c=3}(a)=a/3.
\end{split}
\end{equation}

Where $\Theta(x)$ is the Heaviside function.
The other one loss function, called Normalized Cross-Entropy (NCE) in \cite{Marie}, is defined as

\begin{equation} \label{eq4}
\begin{split}
L_{NCE} = - \frac{1}{N_{tot}}\sum_{pixel} \sum_{i\,classes}{y^c_i w^c}log(\Theta^c [a(x_i)])
\end{split}
\end{equation}

Furthermore, the mapping function $Q^c$ is defined as
\begin{equation} \label{eq5}
\begin{split}
&Q^{c=0}(a)=s(1-a)\\
&Q^{c=1}(a)=a\Theta(1-a)+s(2-a)\Theta(a-2)\\
&P^{c=2}(a)=s(a-1)\Theta(2-a)+(3-a)\Theta(a-2)\\
&P^{c=3}(a)=s(a-2).
\end{split}
\end{equation}
where s is the softplus function,, and $\Theta(x)$ is the Heaviside function. 

Weighted modified and Normalized cross-entropy losses are naturally combined with standard cross-entropy loss and mitigate the class unbalance problem. They also have the ability to encode of any hierarchical and mutually-exclusive topological relationship of classes in a network architecture.

\subsection{Evaluation metrics}
In the task for BRATS, the number of positives and negatives are highly unbalanced. Consequently, four typical different metrics are used by the organizers to evaluate the performance of the algorithm and then rank the different teams.

Give a ground-truth segmentation map G and a segmentation map corresponding one class generated by the algorithm. The four evaluation criteria are defined as following.

Dice similarity coefficient(DSG):
\begin{equation} \label{eq5}
\begin{split}
DSC=\frac{2(G\cap{P})}{|G|+|P|}
\end{split}
\end{equation}

The Dice similarity coefficient measures the overlap in percentage between G and P.

Hausdorff distance (95th percentile) is defined as :
\begin{equation} \label{eq5}
\begin{split}
H(G,P)=max(supinf_{x\in G,y\in P}d(x,y),supinf_{y\in P,x\in G}d(x,y))
\end{split}
\end{equation}
where $d(x,y)$ denotes the distance of x and y, $sup$ denotes the supremum and $inf$ for the infimum. This measures how far two subsets of a metric space are from each other. As used in this challenge, it is modified to obtain a robustified version by using the 95th percentile instead of the maximum(100 percentile) distance.

Sensitivity (also called the true positive rate) measures the proportion of actual positives that are correctly identified. Specificity (also called the true negative rate) measures the proportion of actual negatives that are correctly identified. Assume $P$ is the number of real positive prediction pixel of lesion and $N$ is the number of real negative prediction pixel of lesion. Condition positive $P$ consists with true positive $TP$ and false negative $FN$. Besides, the condition negative $N$ is also divided into $TN$ true negative and $FP$ false positive.

Then, the metrics of Sensitivity and Specificity are illustrated as:
\begin{equation} \label{eq5}
\begin{split}
Sensitivity=\frac{TR}{P}=\frac{TP}{TP+FN}
\end{split}
\end{equation}

\begin{equation} \label{eq5}
\begin{split}
Specificity=\frac{TN}{N}=\frac{TN}{TN+FP}
\end{split}
\end{equation}

Then the values of those four metrics were computed by the organizers independently and made available in the validation leaderboard. 
\section{Experiment results}
\begin{figure}[!htb]
\centering
\includegraphics[width=12cm]{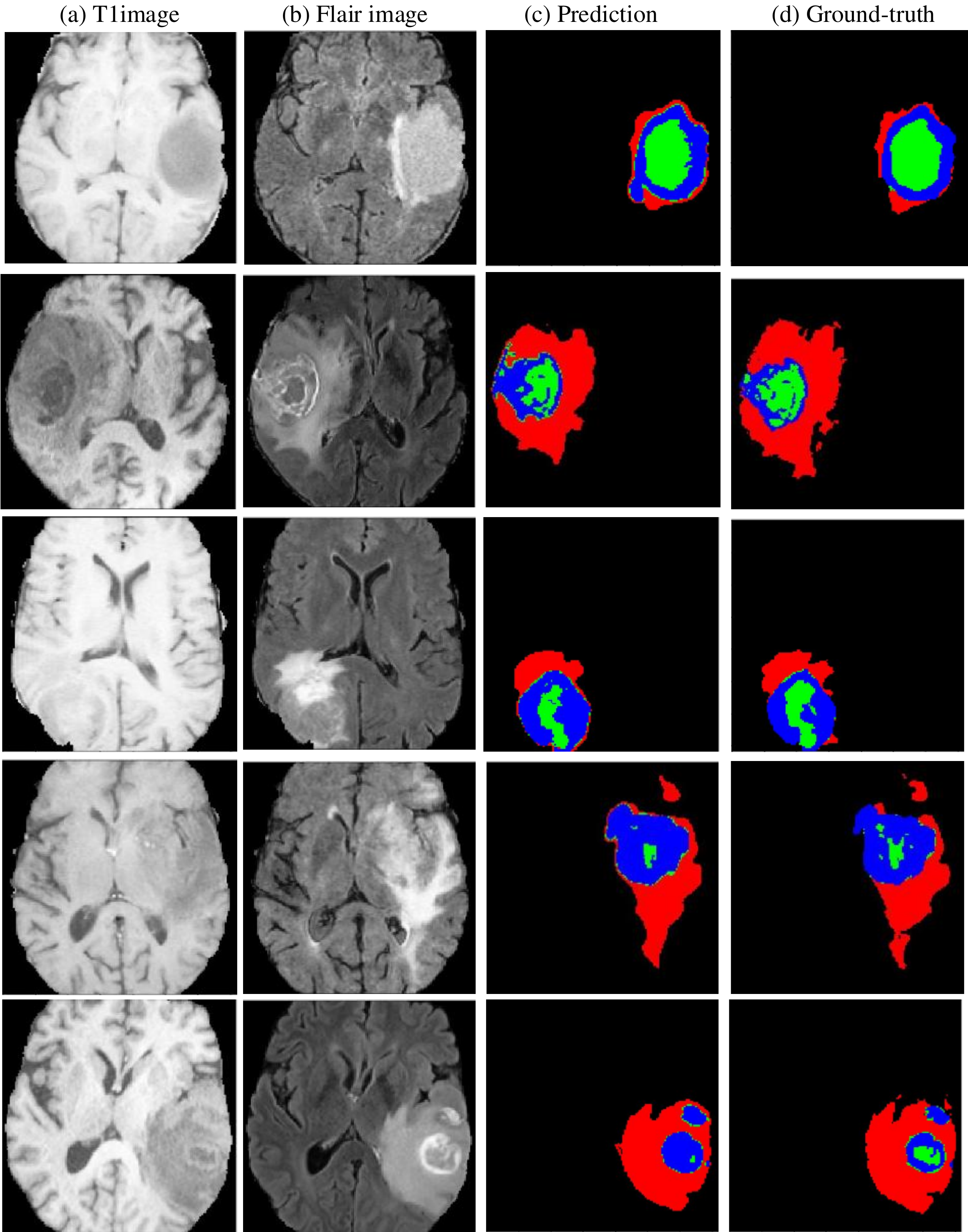}
\caption{Segmentation results, for five different validation cases. The tumor class is depicted in red, tumor core in green and enhancing tumor in blue.  }\label{fig:fig4}
\end{figure}
In BRATS 2018 dataset \cite{Menze,Bakas,Bakas1,Bakas2}, there are four labels, Necrotic core, Edema, Non-enhancing core and Enhancing core that form the three tumor classes in Fig.2. The dataset contains 4 different modalities for MRI, native (T1), post-contrast T1-weighted (T1Gd), T2-weighted (T2) and T2 Fluid Attenuated Inversion Recovery (FLAIR) which are all used as different input channels. We train the networks using ADAM optimizer with an initial learning rate of 0.0005, and to regularize the network, we use early stopping when the precision on the 20\% of the training dataset reserved for validation is no longer improved, and dropout (with rate 0.3) in all residual block before the multi-class sigmoid function. Some slices of segmentation results containing the tumor, tumor core and enhancing core are shown in Fig.4. We observe that the topology geometry between different labels is constrained to the nested-classes relationship, consequently avoiding errors stemming from the lack of topological prior. 

\begin{table*}[h]
\setlength{\abovecaptionskip}{10pt}
\centering
\begin{tabular}{|c|l|c|c|c|c|}
\hline
\multicolumn{2}{|c|}{\multirow{2}{*}{}} & \multicolumn{4}{c|}{Dice score} \\ \cline{3-6} 
\multicolumn{2}{|c|}{}                   &      Enhancing core&      whole tumor&tumor Core & Weight scheme     \\ \hline
\multicolumn{2}{|c|}{ Multi-level(MCE)}                 &       \textbf{0.719}&       0.857&\textbf{0.769} &0.4   \\ \hline
\multicolumn{2}{|c|}{Multi-level(NCE)}                  &       0.676&       0.857&0.755 &0.4     \\ \hline
\multicolumn{2}{|c|}{Multi-level(NCE)}                  &       0.633&       0.837&0.736 &0.5     \\ \hline
\multicolumn{2}{|c|}{Multi-level(NCE)}                  &       0.655&       0.856&0.758 &0.3     \\ \hline
\multicolumn{2}{|c|}{Softmax-based method}                  &       0.691&       \textbf{0.861}&0.763&-      \\ \hline
\end{tabular}
\caption{Validation results presented on the leaderboard}\label{tab:table1}
\end{table*}

The segmentation result is severely affected by highly unbalanced problems existing in the Brats dataset. 
As  class  imbalance  in  a data  set  increases,  the  performance  of  a  neural  net  trained on that data has been shown to decrease dramatically \cite{Mazurowski}. In order to mitigate this issue, many methods \cite{Milletari,Sudre,Crum} were proposed to modify the loss function to alleviate this problems.
Here,the weighted cross entropy incorporating the nested-class information is proposed and investigated. We experimented with different weighting schemes ($\alpha$=1,0.5,0.4,0.3) and with the different losses （MCE and NCE） proposed in \cite{Marie}. The best performing combination turned out to be $\alpha$=0.4 and MCE loss function. The segmentation thresholds to determine the boundaries between classes, were set to [0.95,1.65,2.2] on the validation process. For this final configuration, we reached Dice scores of 86\% for the complete tumor, 77\% for the tumor core and 72\% for the enhancing core as presented in Table 1.
\begin{table}[h]
\setlength{\abovecaptionskip}{10pt}
\centering
\begin{tabular}{|c|c|c|c|}
\hline
 Dice score&  Enhancing core&  whole tumor& tumor Core \\ \hline
 Mean&  0.71965&  0.85685&0.76906  \\ \hline
 StdDev&  0.28526&  0.09802& 0.21962 \\ \hline
 Median&  0.84268&  0.87823&0.84325  \\ \hline
 25quantile&  0.6889&  0.83379&0.70743  \\ \hline
 75quantile&  0.8876&  0.90895&0.91292  \\ \hline
\end{tabular}
\caption{Quantitative evaluation of Dice score }\label{tab:table2}
\end{table}
The weighted-modified-cross-entropy performs much better than the result achieved by normalized cross-entropy, and weight scheme affects the segmentation result severely since the extraordinary unbalance problem. The different weight schemes $[0.5,0.4,0.3]$ are compared and the optimal weight scheme is taken as 0.4. In comparison with the softmax-based method based on the same network architecture proposed by Isensee without ensembles operation, any complicated image pre-processing and post-processing steps and extra training dataset, it indicates that the Dice score of nested-class (enhancing core) drastically improved from 0.691 to 0.719 while the Dice core of whole tumor and tumor core almost remains at same extent. 
\begin{table}[h]
\setlength{\abovecaptionskip}{10pt}
\centering
\begin{tabular}{|c|c|c|c|}
\hline
Mean&  Enhancing core&  whole tumor& tumor Core  \\ \hline
Sensitivity & 0.74119 & 0.93916 &0.78743  \\ \hline
Specificity &  0.9974&  0.98715& 0.99591 \\ \hline
Hausdorff95 &  5.50007&  10.84397&9.98557  \\ \hline
\end{tabular}
\caption{Sensitivity, Specificity and Hausdorff95 results presented on the leaderboard}\label{tab:table2}
\end{table}
The quantitative evaluation (Mean, std, Median, 25\%, 75\% quantile) of Dice score of enhancing core and whole tumor and tumor core are showed in Table 2. And other evaluation metrics (the proportion of actual positives correctly identified---Sensitivity, the proportion of actual negatives correctly identified---Specificity and Hausdorff95) are listed in Table 3. 

\subsection{Threshold scheme definition and analysis }
\begin{figure}[!htb]
\centering
\includegraphics[width=14cm]{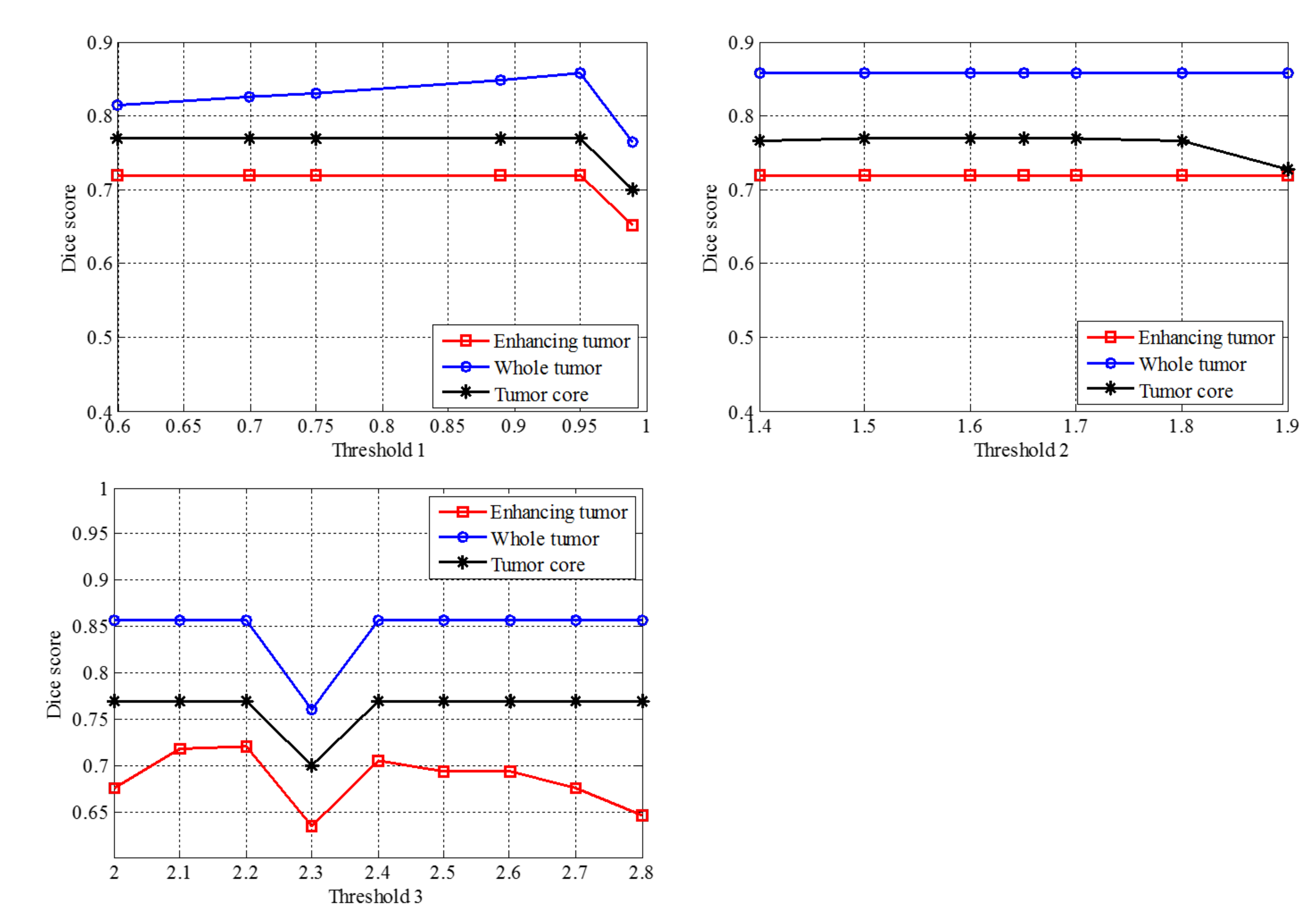}
\caption{Boundary division of Threshold scheme}\label{fig:fig3}
\end{figure}

Setting the optimal threshold is an important component of the multi-class segmentation task, and it is straightforwardly linked  to segmentation boundary. From the activation function (4 nested-class sigmoid function) Fig.3, the 4 classes segmentation problem is corresponding with the threshold scheme with 3 parameters [Threshold-1, Threshold-2, Threshold-3]. The threshold scheme is optimally chosen during the validation procedure, and then fixed and applied into test dataset.

In order to analyze how the threshold affects the segmentation accuracy, the relationship between boundary threshold and Dice score is illustrated in Fig.5.  The target threshold is changed to the value taken from a specific interval which is considered to be possible to achieve optimal segmentation result when other thresholds are fixed at the optimal value.  The criteria Dice score of three classes is very sensitive to the threshold-3 value compared with other two threshold indexes, that it may drop into Dice score valley within interval [2.2,2.4]. The threshold-2 index has little impact on the Dice score of whole classes except for threshold greater than 1.8. Consequently, it is easier to make an optimal threshold scheme after determining indexes of threshold-3 and threshold-2. After experiment and optimization, the suitable threshold scheme in the Brats challenge is selected as [0.95,1.65,2.2].

\section{Conclusions}
In this paper we applied the technique of multi-level activation to the nested classes segmentation of glioma. The results of our experiments indicate that the multi-level activation function and its corresponding loss function are efficient compared to Softmax output layer based on the same network framework. Using the MCE loss function and a reweighting scheme with power-law =0.4, we obtain Dice scores 86\% for complete tumor, 77\% for tumor core and 72\% for enhancing core on the validation leaderboard of the 2018 BRATS challenge, proving the applicability of the multi-level activation scheme. Finally, this activation could be combined with other network architectures. Using it with the best performing architecture of the BRATS challenge could even lead to further improved results.
%
%
%

\end{document}